\documentclass[conference]{IEEEtran}
\pdfoutput=1  
\IEEEoverridecommandlockouts
\usepackage{cite}
\usepackage{amsmath,amssymb,amsfonts,mathrsfs}
\usepackage{graphicx}
\usepackage{textcomp}
\usepackage{xcolor}
\usepackage{mathtools}
\usepackage{float}
\usepackage{url}
\usepackage{dblfloatfix}

\DeclareMathOperator*{\argmin}{argmin}
\def\BibTeX{{\rm B\kern-.05em{\sc i\kern-.025em b}\kern-.08em
    T\kern-.1667em\lower.7ex\hbox{E}\kern-.125emX}}
\usepackage[hidelinks,breaklinks]{hyperref}
\usepackage{eso-pic}

\begin{document}

\title{HorizonNet for visual terrain navigation}

\author{\IEEEauthorblockN{Bertil Grelsson*$^{1}$, Andreas Robinson*$^{1}$, Michael Felsberg$^{1}$, Fahad Shahbaz Khan$^{1,2}$}
\IEEEauthorblockA{\textit{$^{1}$Computer Vision Laboratory, Department of Electrical Engineering, Link\"oping University, Sweden} \\
\textit{$^{2}$Inception Institute of Artificial Intelligence, Abu Dhabi, UAE}\\
\{bertil.grelsson, andreas.robinson, michael.felsberg, fahad.khan\}@liu.se}

\thanks{*These authors contributed equally.}
}

\maketitle
\AddToShipoutPictureBG*{%
  \AtPageLowerLeft{\hspace*{49.15pt}\raisebox{20pt}{%
    \parbox{516pt}{\scriptsize
      \copyright~2018 IEEE. Personal use of this material is permitted. Permission
      from IEEE must be obtained for all other uses, in any current or future media,
      including reprinting/republishing this material for advertising or promotional
      purposes, creating new collective works, for resale or redistribution to servers
      or lists, or reuse of any copyrighted component of this work in other works.
      Published in \emph{2018 IEEE International Conference on Image Processing,
      Applications and Systems (IPAS)}, pp.~149--155.
      DOI: \href{https://doi.org/10.1109/IPAS.2018.8708868}{10.1109/IPAS.2018.8708868}\par}}}%
}

\begin{abstract}
This paper investigates the problem of position estimation of unmanned surface vessels (USVs) operating in coastal areas or in the archipelago.
We propose a position estimation method where the horizon line is extracted in a 360 degree panoramic image around the USV.
We design a CNN architecture to determine an approximate horizon line in the image and implicitly determine the camera orientation (the pitch and roll angles). The panoramic image is warped to compensate for the camera orientation and to generate an image from an approximately level camera. A second CNN architecture is designed to extract the pixelwise horizon line in the warped image. The extracted horizon line is correlated with digital elevation model (DEM) data in the Fourier domain using a MOSSE correlation filter. Finally, we determine the location of the maximum correlation score over the search area to estimate the position of the USV. Comprehensive experiments are performed in a field trial in the archipelago. Our approach provides promising results by achieving position estimates with GPS-level accuracy.
\end{abstract}


\section{Introduction}
\label{sec:intro}
In recent years, unmanned systems such as Unmanned Aerial Vehicles (UAVs), Unmanned Ground Vehicles (UGVs), and Unmanned Surface Vessels (USVs) have become increasingly popular providing safe and secure operations in remote environments.
Within unmanned systems, the aim of USVs is to perform various ocean sensing tasks in a variety of cluttered sea environments. Generally, these USVs (autonomous or tele-operated) are reliant on accurate position measurements provided by the Global Positioning System (GPS) for safe navigation. In hostile ocean scenarios, the GPS signal can be unreliable or not available at all, creating the need for alternative position sensors. In this paper, we tackle the problem of providing accurate position measurements for a USV operating in challenging coastal areas or in the archipelago.

\begin{figure}[t]
  \begin{center}
  \includegraphics[width=0.560\linewidth]
                   {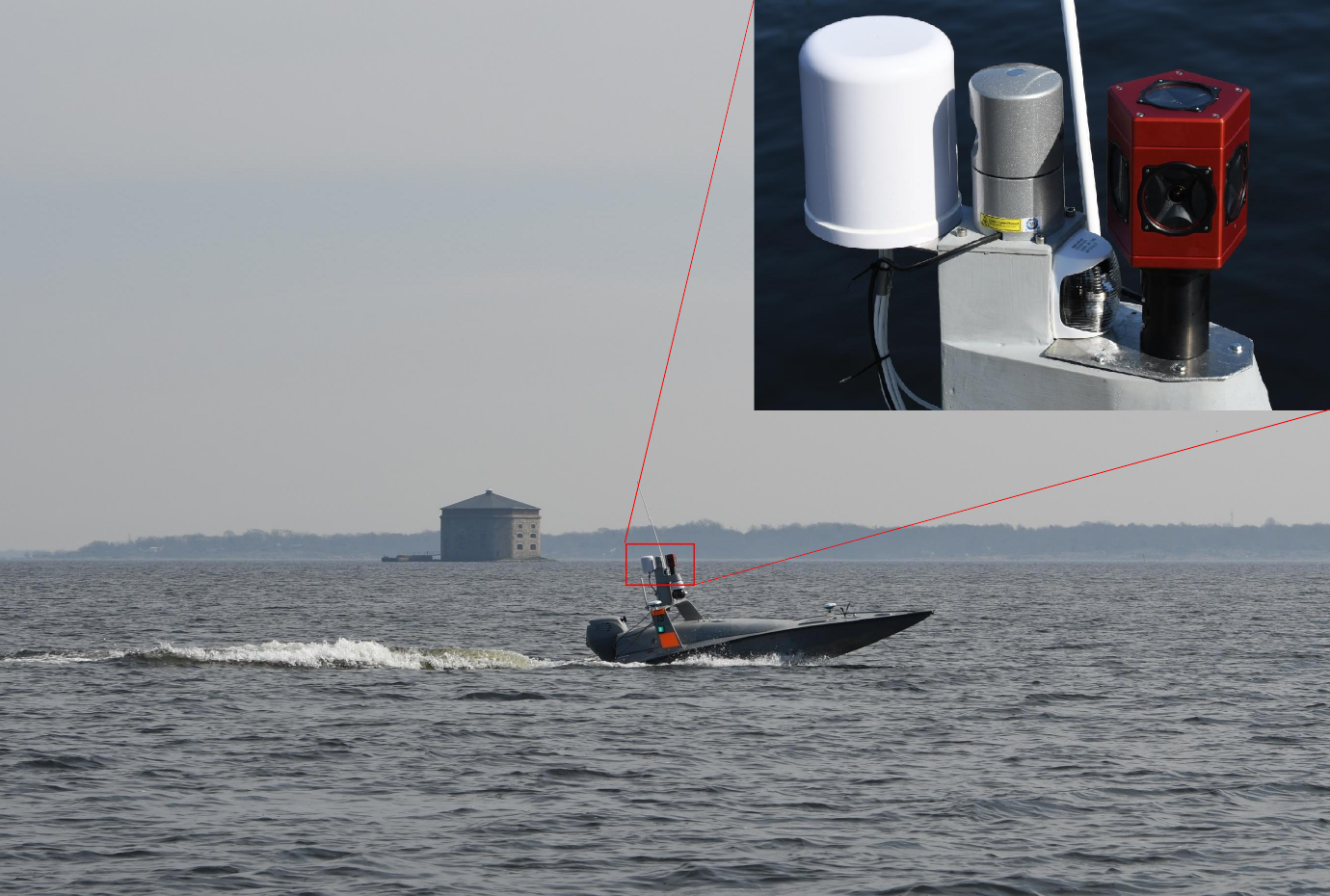} 
  \includegraphics[width=0.420\linewidth]
                   {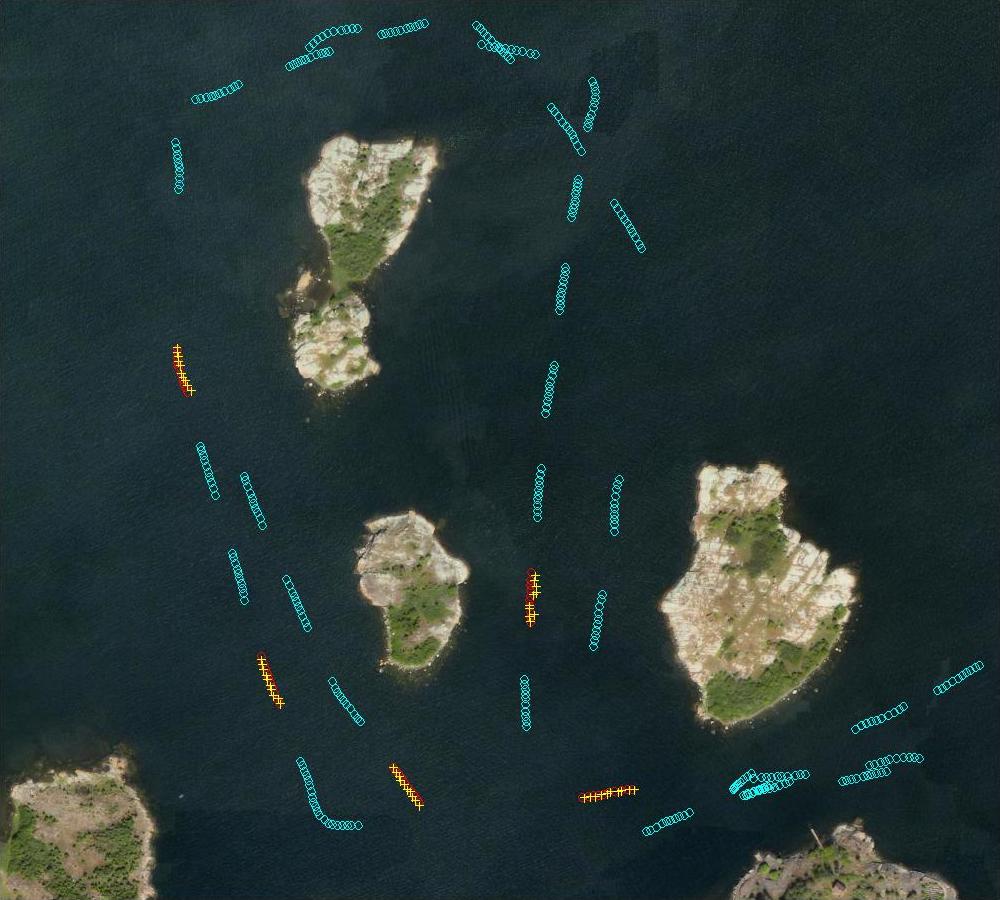}
  \end{center}
  \vspace{-3mm}
  \caption{Left: Image showing the USV equipped with a Ladybug omnidirectional camera. Right: CNN training image positions (cyan); Test image position estimation with the GPS (red) and the proposed method (yellow). Our proposed method provides position estimation as accurate as the GPS.}
  \vspace{-5mm}
  \label{fig:USV_position_estimates}
\end{figure}

Terrain navigation, without any need for externally controlled signals and sensors, is suitable as an alternative position sensor in these scenarios.
In terrain navigation, measurements of the surrounding terrain are correlated with a terrain spatial database to provide position measurements.
Terrain navigation is well established as a position sensor and the underlying sensor measurements could originate from various techniques such as radars, sonars, altimeters, lidars, and cameras \cite{vaman2012trn,han2016gps,melo2017survey}.
For the position estimation in our scenario, the USV can utilize information obtained from an omnidirectional camera, a digital compass, and a high-resolution digital elevation model (DEM) of the operational area, see Figure \ref{fig:USV_position_estimates}.

In recent years, deep learning and convolutional neural networks (CNNs) have significantly boosted the level of performance for various computer vision tasks including image classification, object segmentation, and object tracking.
For unmanned systems, CNNs have e.g. been used for object recognition in UAV images \cite{radovic2017object}, road lane guidance for autonomous cars \cite{nugraha2017towards}, and collision avoidance manoeuvers for USVs \cite{xu2017deep}.
In this paper, we employ CNNs to extract terrain information from USV images to generate accurate position measurements for the USV.

To aggregate terrain information from consecutive images and to align the image content with DEM data for USV position estimation, object tracking and registration are essential components. In recent years, the best performing visual object trackers apply a discriminatively trained correlation filter (DCF) on top of multidimensional features \cite{diva2:1192158}. The foundation for DCF-based trackers is the MOSSE correlation filter \cite{bolme2010visual}. A filter is trained to model the appearance of the tracked object in some example images. To search for the object in the next frame or, in our case, to register the object with DEM data, a correlation score is computed over the search area. For computational speed, the correlation is performed in the Fourier domain.

For terrain navigation of a manned surface vessel in coastal areas and in the archipelago, humans would intuitively use readily observable characteristic landmarks in a few directions, project the directions on a sea chart or map, and use cross bearing to determine the vessel position.
The horizon line most often constitutes a spatially extended characteristic landmark for islands and the shore. Matching of the complete horizon line around the vessel with a map is mathematically a more robust cross bearing measurement than just using a few directions. To obtain highly accurate position estimates, the horizon line must be captured with high angular resolution, which camera sensors can provide. These insights are the motivations behind our proposed position estimation method.

We propose a position estimation method where two CNNs are employed to extract the camera orientation and horizon line, respectively, in a 360$^{\circ}$ panoramic image around the USV.
The horizon line is correlated with DEM data in the Fourier domain using a MOSSE correlation filter.
Finally, we determine the location of the maximum correlation score over the search area to estimate the position of the USV.
Our approach is evaluated in a field trial in the archipelago and achieves promising results with the average global position accuracy of 2.5$\pm$1.3 meters relative to the GPS ground truth data, i.e. we achieve position estimates with GPS-level accuracy.
Figure \ref{fig:USV_position_estimates} shows position estimates with GPS and the proposed method.

Our contributions are: 1) The proposed method for USV terrain navigation, 2) CNNs designed for camera orientation estimation and horizon segmentation in a marine environment, 3) Horizon line registration with a MOSSE correlation filter, 4) Comprehensive experiments that demonstrate the GPS-level accuracy of the proposed method.

\section{Related work}
\label{sec:related_work}

Automatic extraction of the horizon line and water line (which is the first water to land/sky transition seen from the vessel) in the image is equivalent to segmentation of the image into sky/land/water groups. Previous works \cite{lee2017semantic,verbickas2014sky} have investigated the use of CNNs to determine straight lines and sky segmentation in images. Lee et al. \cite{lee2017semantic} propose an approach where a CNN is trained to find straight lines in order to extract semantically meaningful information from the image. Verbickas and Whitehead \cite{verbickas2014sky} propose a CNN-based method for sky segmentation. In their approach a two convolutional layer network is trained from scratch on the authors' own dataset.
A comparison of deep learning methods for horizon/sky line segmentation is presented by Ahmad et al. \cite{ahmad2017comparison}. However, their evaluation is only performed on sky/mountain images. To the best of our knowledge, we have not encountered any public benchmark or CNN-based segmentation method with respect to marine images and USVs.

Most previous work on horizon registration for position estimation has been performed in the spatial domain.
Dumble and Gibbens \cite{dumble2015efficient} precompute reference horizon profiles from DEM data in the Alps in a set of 3D grid points. They extract the reference profile at the grid point closest to the assumed position of their aerial vehicle. To refine the estimated location, they use gradient descent to iteratively minimize the error between the horizon line in the image and the transformed reference profile.
The method requires a large horizon profile variance, which prevents its use in our scenario in the archipelago with low altitude islands.
A method for accurate registration of low variance horizon profiles is proposed by Grelsson et al. \cite{grelsson2016highly}, but they only estimate the camera orientation and not the position.
Another method suitable for low variance horizon lines is proposed by Chiodini et al. \cite{chiodini2017mars}, where a Mars rover is localized by matching the detected skyline with DEM data. For position estimation, they do a grid search over the location and the viewing angle, and minimize the least-square error between the detected and the rendered skyline.
Baatz et al. \cite{baatz2012large} use a database with extracted features from the horizon line called contourlets, and employ a Bag-of-Words-like approach for large scale localization of mountainous images.
We have not encountered any previous work where registration of the horizon line with DEM data has been carried out in the Fourier domain.

The introduction of the MOSSE correlation filter \cite{bolme2010visual} showed that image object tracking with adaptive correlation filters in the Fourier domain is significantly faster and also more robust to variations in target appearance than previous trackers working in the spatial domain.
A major advantage with the proposed method compared to appearance based approaches \cite{arandjelovic2016netvlad} is that it does not require images to have been captured in the test area prior to navigation. It is sufficient to have images from a similar environment to provide accurate position estimates.

\section{Classical methods for position estimation}
\label{sec:classical_position_estimation_methods}
\begin{figure}[t]
  \begin{center}
  \includegraphics[width=0.98\linewidth] 
                   {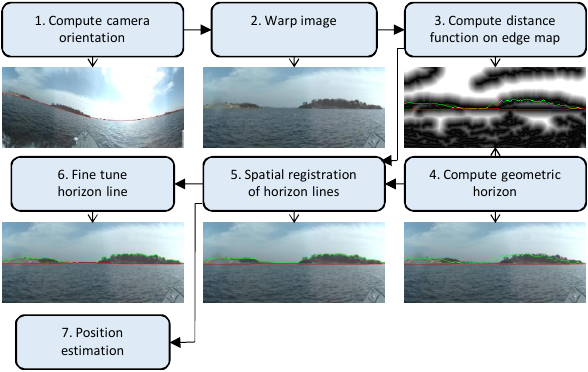}
  \end{center}
  \vspace{-3mm}
  \caption{Flowchart for position estimation with classical methods.
  Predicted horizon line (red) overlaid on image (1). Geometric horizon line (green) and water line (red) overlaid on distance function (3) and warped image (4) before registration, after registration (5), and after fine tuning (6).}
  \vspace{-5mm}
  \label{fig:flow_chart_classical_algorithm}
\end{figure}

The approach for our position estimation method is based on registration of the horizon line with DEM data. In this section, we sketch an algorithm with classical computer vision methods for position estimation. This algorithm is used to create the ground truth labels for the CNNs in the proposed method.
It also provides a baseline for the position estimate accuracy that can be obtained by registration in the spatial domain. An algorithm flowchart is shown in Figure \ref{fig:flow_chart_classical_algorithm} together with cropped sample images for illustration of each step.

The camera used in our field trials generates panoramic images in a cylindrical projection.
First, to determine an approximate camera orientation from the image, we use Canny edge detection \cite{Canny} and Hough voting \cite{Hough}. We search for the approximate horizon plane on the cylinder, which will be an S-shaped curve in the panoramic view.
We adapt the method in Grelsson et al. \cite{grelsson2016highly} and vote for the normal vector of the horizon plane parameterized with the pitch and roll angles.

The second step in the algorithm is to warp the image to compensate for the camera orientation. The warping will create an image corresponding to an approximately level camera, suitable for the subsequent registration process with the geometric horizon.
%
%
%
%
%
%
%
As a third step, we compute a Canny edge image on the warped image and a distance function $\mathrm{D}$ based on the edge image.
Another input to the registration is the geometric horizon line from the digital elevation model. The DEM for the test site was provided by Vricon
\cite{Vricon}.
The DEM is computed from recent satellite imagery and has a pixel resolution on the ground of 0.5m. The altitude accuracy is in the same order as the pixel resolution.

To create the ground truth labels for training the CNNs, we use the position given by the GPS and the heading angle from the digital compass or, if not available, the tangent vector from the logged GPS trajectory.
The geometric horizon is generated by ray-tracing using DEM data.
For the desired number of image columns around the vessel, we extract the altitude profile from the DEM. In each direction, we compute the elevation angle of all objects along the ray.
The maximum elevation angle is taken as the horizon point in that direction. The water line point, the first water to land/sky transition in each direction, is taken as the vertical viewing angle to the point where the DEM makes a step larger than a 0.2m threshold along the ray direction. The search radius is set to 6km in all directions.
The ideal horizon, assuming a flat and spherical earth, is at 3.2km for a camera at 1.0m height, but we add some margin to cope with the topography.

Ideally, the complete geometric horizon line would be located where the distance function $\mathrm{D}$ is zero. To find the horizon line in the image, we search for a rotation of the geometric horizon line points on the cylinder, such that when projected onto the image, their summed distance function values will be a minimum, i.e. we minimize the score
\begin{equation}
  s = \argmin_{\theta,\phi,\psi} \sum_i \mathrm{D}\left\{\mathbf{R}(\psi)\pi(\mathbf{R}(\theta,\phi)\pi^{-1}(h_i))\right\}  
\label{eq:spatial_position_estimation}
\end{equation}
where $h_i$ are the horizon line points, $\pi$ is the projection from the cylinder to the image plane, $\theta$, $\phi$, and $\psi$ are the pitch, roll and heading angles and $\mathbf{R}$ is a rotation matrix.

For registration, we perform a grid search over the pitch, roll and heading angles. For the first two angles we need to compute the rotation matrix and project the transformed points onto the image plane. The heading angle rotation simply corresponds to a horizontal shift on the image plane. The step size in pitch and roll is set to 0.25$^{\circ}$ and we search over the range $\pm$2$^{\circ}$.
We extract the rotation angles for the minimum score and project the geometric horizon line and water line onto the warped image after transformation with the said rotation angles.
In general, there is a good fit between the geometric horizon line and water line with the image content, but occasionally there are small deviations. The main reason is that the DEM is not a perfect representation of the real world. To adjust for the discrepancies we perform a final tuning step of the geometric horizon line and water line in the image.

For final tuning we first convolve the warped image with a Sobel filter to enhance gradients in the vertical direction. For each image column we search for a local maximum of the gradient, exceeding a threshold, in a small region close to the horizon and water lines obtained in the registration step.
If no gradient maximum is detected in the search region, the horizon line from the registration is retained.
For position estimation, the algorithm is run for positions over a grid and the minimum score obtained according to (\ref{eq:spatial_position_estimation}) is recorded for each position.

The main drawback with this classical method for position estimation is that it is prohibitively slow for real-time applications. In our implementation, the spatial registration took more than 1s per position grid point, whereas our proposed method below achieved more than 40 grid points per second.


%

\section{CNN-based position estimation method}
\label{sec:position_estimation_method}
Prior to designing our proposed method, we experimented with an end-to-end CNN to output the position estimate directly from the input image. We tried a modified version of Posenet \cite{kendall2015posenet}. We trained it to learn the camera position and orientation, but the training failed completely. The reason for this behavior is quite simple. Only a small part of the image (the land objects) contain information that is useful for position estimation. The sky and sea change appearance over time and will only distract the position estimation if using the full image content. Hence, prior knowledge about what is the relevant part of the image for position estimation is required.

\begin{figure}[t]
  \begin{center}
  \includegraphics[width=0.98\linewidth] 
                   {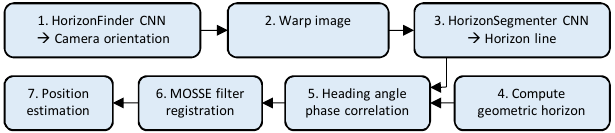}
  \end{center}
  \vspace{-3mm}
  \caption{Flowchart for position estimation with the proposed method.}
  \vspace{-5mm}
  \label{fig:flow_chart_proposed_algorithm}
\end{figure}

This insight motivates the design of our proposed method for position estimation, which is similar to the classical method in its architecture.
The proposed algorithm consists of seven steps, which are explained in detail in this section and shown in a flow chart in Figure \ref{fig:flow_chart_proposed_algorithm}.

\noindent\textbf{HorizonFinder CNN.}
To determine a rough camera orientation (pitch and roll angles), i.e. to find an approximate horizon line in the panoramic image, we employ a convolutional neural network called HorizonFinder. This CNN replaces the Canny detector and the Hough voting in the classical method.

\begin{figure*}[t]
  \vspace{-3mm}
  \begin{center}
  \includegraphics[width=0.90\linewidth] 
                   {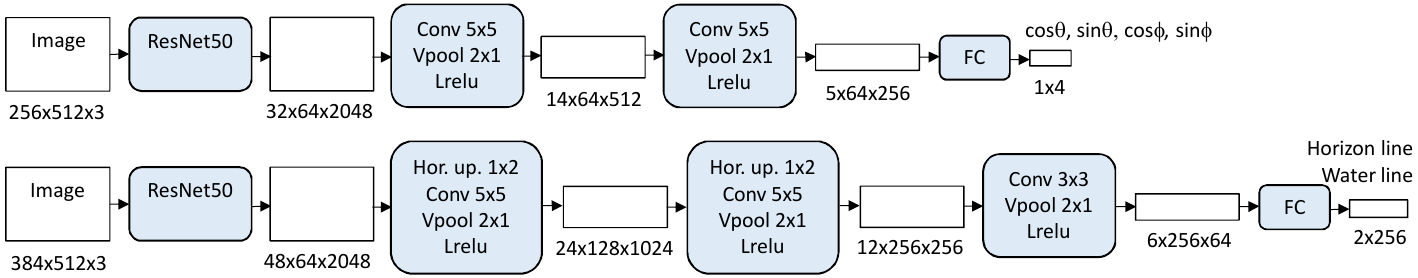}
  \end{center}
  \vspace{-5mm}
  \caption{Network design for HorizonFinder (top) and HorizonSegmenter (bottom). The numbers denote the input and output size (H$\times$W$\times$C) for each layer.}
  \label{fig:CNN_network_design}
\end{figure*}

We use a ResNet50 network \cite{he2016deep} pretrained on ImageNet to provide feature descriptors. We then add two stacks with a convolutional layer, vertical pooling, and a Leaky Rectified Linear Unit (Lrelu) \cite{maas2013rectifier} activation function. Finally, we have a fully connected layer to output the camera orientation angles. We found that the network training was more accurate when the output was the cosine and sine of the  pitch and roll angles, $\theta$ and $\phi$, rather than having the attitude angles by themselves as output. The network design is shown in Figure \ref{fig:CNN_network_design}.

As a loss function we use the L1 loss between the predicted and ground truth values for the parameters $\cos\theta$, $\sin\theta$, $\cos\phi$, and $\sin\phi$. This network design and loss function gave satisfactory results as judged by viewing the backprojected horizon line on the panoramic image.
Since we did not have access to any exact ground truth for the camera orientation from external sensors, we could not perform any quantitative comparison for different network designs and loss functions.
We trained the network for 100 epochs. The initial learning rate was set to 0.0001 and it was then reduced by a factor of two every nine epochs.


\noindent\textbf{Image warping.}
Based on the predicted pitch and roll angles from the HorizonFinder CNN, we warp the panoramic image to compensate for the camera orientation.
Since the horizon now will be almost vertically centered in the image, we only warp the central part of the image.
The size of the original panoramic image is 2048$\times$1024, whereas the warped image is 2048$\times$384.

\noindent\textbf{HorizonSegmenter CNN.}
To predict the location of the horizon line and water line in the warped image, we design a second CNN called HorizonSegmenter. This CNN has no one-to-one counterpart in the classical method.
We use a similar network design as for HorizonFinder. We start with a pretrained ResNet50 to generate feature descriptors. We use three stacks of layers, each comprising a convolutional layer, vertical pooling, and a Lrelu activation function. To obtain the same horizontal resolution as the ground truth, we insert two horizontal upsampling layers with bilinear interpolation. Finally, a fully connected layer is added to output the vertical pixel location of the horizon line and water line for all image columns. As a loss function we use the absolute pixel difference between the predicted and ground truth horizon line and water line summed over the training image.
The network design is shown in Figure \ref{fig:CNN_network_design}.
To avoid considerable overfitting during training due to relatively few images, we randomly generate training images from the original warped images. We make a random horizontal crop of a 512$\times$384 image from the original image during training.
We used the same learning rate scheme as for the HorizonFinder network.

\noindent\textbf{Compute geometric horizon.}
The geometric horizon is computed in exactly the same manner as described in the classical position estimation method.

\noindent\textbf{Phase correlation for relative heading angle measurements.}
For the registration with the MOSSE correlation filter in the next step, we need the camera heading angle for each image in a global coordinate system as an input. The absolute heading angle of the first image in a sequence can be obtained from a digital compass (without relying on GPS) with an accuracy better than 2$^{\circ}$ \cite{Airmar_digital_compass}. In our field trial we did not have access to a digital compass. Instead, we replaced the heading for the first image with the ground truth heading from the GPS trajectory plus some noise simulating the digital compass.

To find the relative change in heading angles from one video frame to the next, we employ phase correlation matching \cite{meneghetti2015image} of the predicted horizon line output from the HorizonSegmenter network for successive images.

%

\noindent\textbf{MOSSE correlation filter.}
We adapt the MOSSE correlation filter \cite{bolme2010visual}, originally developed for visual object tracking. The MOSSE filter is designed to generate a desired output signal (typically a Gaussian) shifted to the spatial location most closely corresponding to a set of learned reference signals. For efficiency, the filter is evaluated in the Fourier domain. The filter is trained with multiple references to improve its robustness against changes in appearance and noise.

In our case we use the segmented horizon lines from an image sequence as reference.
We align the segmented horizon lines in the spatial domain in accordance with their estimated heading angle. In the Fourier transform domain, we denote the horizon lines in the sequence with $F_i$ and the Gaussian target signal with $G$.
We compute the MOSSE filter $K$ as
\begin{equation}
  K^\star = \frac{\sum_i G_i \cdot F_i^\star}{\sum_i F_i \cdot F_i^\star} \enspace ,
\label{eq:mosse_filter_horizon}
\end{equation}
and the MOSSE filter signal response as
\begin{equation}
  r = \mathscr{F}^{-1} \left\{H \cdot K^\star \right\} \enspace ,
\label{eq:mosse_filter_signal_response}
\end{equation}
where $H$ is the Fourier transform of the geometric horizon line from the DEM, $^\star$ is the complex conjugate and $\cdot$ denotes element-wise multiplication.
Ideally, $r$ will be the Gaussian target signal with a zero phase shift if the heading angle estimate is correct.
Since the digital compass gives the absolute heading with an accuracy around 2$^{\circ}$, we search for the peak signal response within a $\pm$4$^{\circ}$ band from the center to have some margin.
As a quality measure of the signal response, i.e. our MOSSE filter correlation score, we use the peak-to-output-energy ratio \cite{javidi1994design}.
We suppress the response within the expected Gauss signal bandwidth around the detected peak and compute the average energy over the remainder of the signal response. The score is the peak signal over the square root of the average energy.

\noindent\textbf{Position estimation.}
To generate a position estimate, we compute the MOSSE filter correlation score for various positions in a XY grid. We extract the maximum score over the search area and apply a second order polynomial fit to obtain a subgrid position estimate.

\begin{figure*}[t]
  \begin{center}

  \includegraphics[width=0.85\linewidth] 
                   {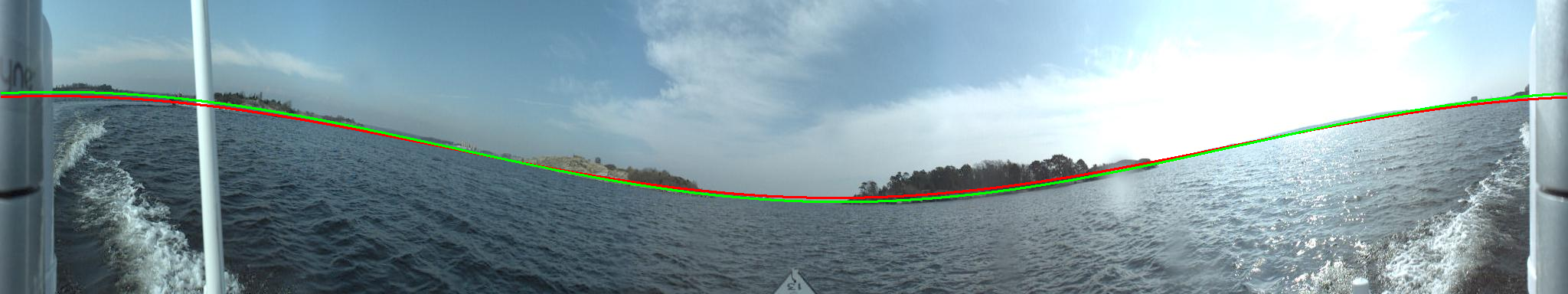} \\
  \includegraphics[width=0.85\linewidth] 
                   {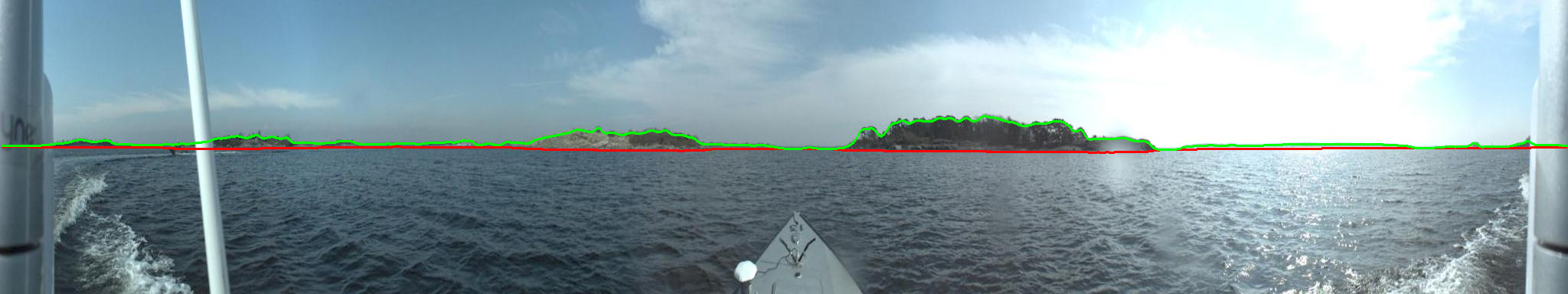}
  \end{center}
  \vspace{-3mm}
  \caption{HorizonFinder result (top): CNN output (green) and ground truth (red). HorizonSegmenter result (bottom): predicted horizon line (green) and water line (red). HorizonFinder predicts the approximate horizon line well and HorizonSegmenter outputs the essential shape of the horizon line.}
  \vspace{-3mm}
  \label{fig:horizon_finder_segmenter_results}
\end{figure*}

\section{Experiments and Results}
\label{sec:experiments_results}

\subsection{Field trial}
\label{subsec:field_trial}
A field trial was performed on a single day in the archipelago. Omnidirectional images were captured with a Ladybug3 camera \cite{ladybug3} mounted on a tele-operated USV (4m long), see Figure \ref{fig:USV_position_estimates}.
A total of 20k images were captured at 10 fps during a 45 minute trial. The rear view (roughly 30$^{\circ}$) was occluded by other sensors on the vessel.
The vessel position was measured with a U-blox EVK-8 GPS receiver \cite{ublox_gps} acquired at 1 fps.


\subsection{CNNs for horizon detection and segmentation}
\label{subsec:CNN_horizon_detection_segmentation}
40 image sequences, each containing 100 consecutive images acquired at 10 fps, were selected for evaluation of the CNNs and the full position estimation method.
Ground truth labels were generated with the method described in Section \ref{sec:classical_position_estimation_methods}.
35 image sequences were used for training of the two CNNs and 5 sequences were used for testing, see Figure \ref{fig:USV_position_estimates}.

\noindent\textbf{HorizonFinder.}
The HorizonFinder CNN robustly locates the horizon line in the panoramic image. The test error relative to the ground truth is on average less than 0.1$^{\circ}$.
Figure \ref{fig:horizon_finder_segmenter_results} shows the result for one test image.

One observation that was made is that the HorizonFinder network tends to give a visually better approximation of the horizon line than the ground truth generated with Canny detection and Hough voting. The reason is probably that the network has learned to generalize and average the errors produced in the ground truth generation.
We have no ground truth camera orientation from other sensors to provide a quantitative evidence for this observation.

\noindent\textbf{HorizonSegmenter.}
The HorizonSegmenter network segments the horizon line and water line well in the warped image, but the segmentation is not perfect.
Figure \ref{fig:horizon_finder_segmenter_results} shows the result for one test image.
The CNN learns to predict the general shape of the horizon line but loses the high frequency variations in the horizon line. Single trees and other thin structures in the true horizon line are missed. This observation must be kept in mind when using the segmented horizon line in the registration process for position estimation, which is our main goal.

\subsection{Position estimates}
\label{subsec:position_estimates}
For position estimation we have evaluated the results achieved with the proposed method (Section \ref{sec:position_estimation_method}) and make a comparison with baseline results obtained with registration in the spatial domain (Section \ref{sec:classical_position_estimation_methods}).

\begin{figure*}[htbp]
  \vspace{-0mm}
  \begin{center}
  \includegraphics[width=0.30\linewidth] 
                   {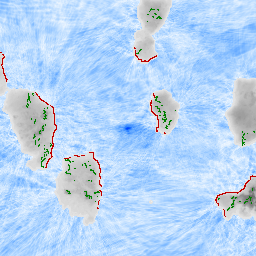}
  \includegraphics[width=0.30\linewidth] 
                   {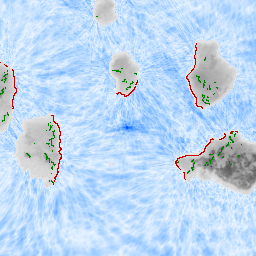}
  \includegraphics[width=0.30\linewidth]
                   {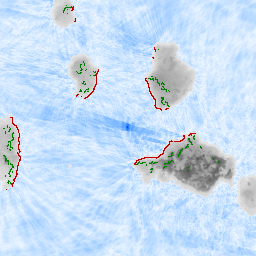}
  \end{center}
  \vspace{-2mm}
  \caption{The MOSSE filter correlation score (blue) in the search region for three example positions. The geometric horizon lines are displayed in green and the water lines in red (as seen from the true position). The search region is extended to 1km$\times$1km for better illustration. The highest correlation score is clearly obtained for the true position.}
  \label{fig:MOSSE_filter_correlation_score}
\end{figure*}

We first assume that we have gained the prior knowledge that the vessel is located within a 80m$\times$80m region. We compute the MOSSE filter correlation score every 4m within this region. We extract the grid position of the maximum score and perform a second order polynomial fit of the scoring function to interpolate a sub-grid resolution position estimate.
To compute the MOSSE filter, we use the segmented horizon line for 10 consecutive images and correlate it with the geometric horizon line computed in one of the grid points.
The USV travels about 6m during these 10 frames (1s), i.e. slightly more than the grid size. Typical results for the correlation score are illustrated in Figure \ref{fig:MOSSE_filter_correlation_score}. 

\begin{figure*}[t]
  \vspace{-2mm}
  \begin{center}
  \includegraphics[width=0.49\linewidth]
                   {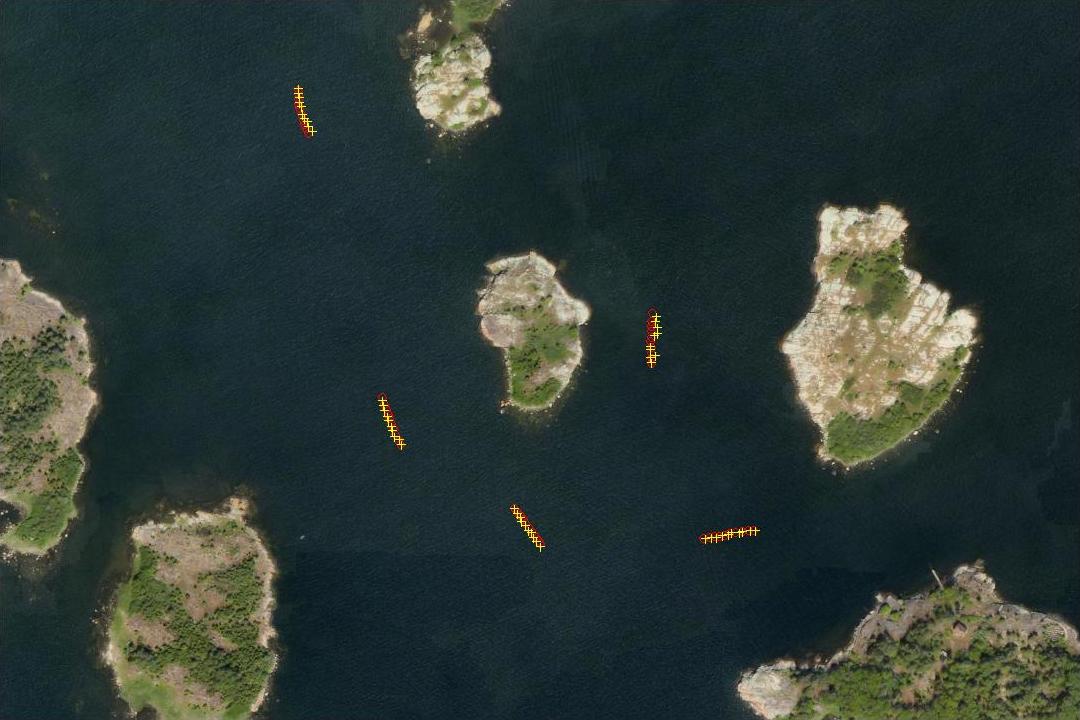}
  \includegraphics[width=0.49\linewidth]
                  {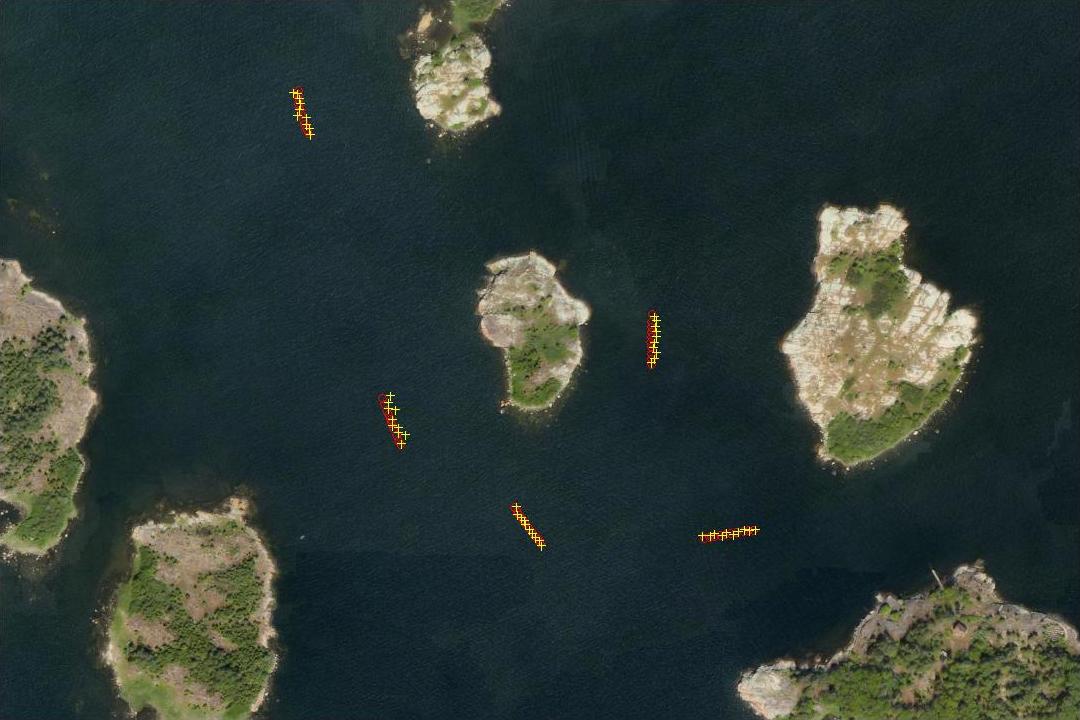} 
  \end{center}
  \vspace{-3mm}
  \caption{Position estimation results: GPS (red), proposed method (yellow). MOSSE filter registration (left) and spatial registration (right).}
  \label{fig:position_estimate_results}
\end{figure*}

The position estimates obtained with the proposed method
 are listed in Table \ref{table:position_estimate_results} and shown on the map in Figure \ref{fig:position_estimate_results} together with the measured GPS position.
The estimated position is the average position for the USV while capturing the 10 images.
The average deviation between the position estimates with our proposed method and the GPS measurements is 2.47$\pm$1.26m.
Assuming uncorrelated position measurements, these deviations imply that our method provides position measurements with GPS-level accuracy, according to e.g.  \cite{gps_accuracy}.


\begin{table}[b]
\caption{Position estimation accuracy achieved with the MOSSE filter registration and with the spatial registration.}
\centering
\begin{tabular}{|c|c|c|c|}
\hline
Registration method & Input & Mean error (m) & Std (m) \\ \hline
MOSSE &  Horizon & \bf{2.47} & \bf{1.26} \\
Spatial, full search &  Horizon & 3.09 & 1.91\\
Spatial, shift only &  Horizon & 6.45 & 5.95 \\
MOSSE &  Horizon+Water & 3.02 & 1.40 \\ \hline
\end{tabular}
\vspace{-0mm}
\label{table:position_estimate_results}
\end{table}

The key contribution in our proposed method is the adaptation of the MOSSE filter to perform the registration of the horizon line in the Fourier domain.
As a comparison, we perform the registration in the spatial domain, where we use the method described in Section \ref{sec:classical_position_estimation_methods}.
For each position grid point, we record the lowest summed distance according to (\ref{eq:spatial_position_estimation}).
We detect the location of the lowest distance over the grid and make the same sub-grid resolution interpolation as for the proposed method.
The average position error obtained with spatial registration is 3.1m.
A t-test at 95\% significance level shows that the position errors achieved with the proposed method are significantly lower than those obtained with spatial registration.
The Fourier method is more robust against deviations between the  horizon line from the image and the geometric horizon line from the DEM. As noted, the segmented horizon line from the CNN does not catch all high frequency variations in the image. Furthermore, not even a high-resolution DEM will be a perfect representation of the real world. These discrepancies are more or less inevitable, and for the Fourier registration, the lower frequency content is sufficient to produce highly accurate position estimates whereas the discrepancies distract the spatial registration more.

The spatial registration is also by far slower than registration in the Fourier domain. To speed up the spatial registration considerably, we searched only in the heading angle (a shift in the image domain), and assumed that the pitch and roll angle estimates from the Hough voting were correct. This speed-up degrades the average position error to 6.5m. Again, this shows that the spatial registration is more sensitive to angular errors than MOSSE registration, which handles these angular errors in the Fourier domain matching process.

The HorizonSegmenter CNN additionally extracts the water line from the image. The motivation is that we wanted to investigate if the water line could aid and improve the position estimates.
We compute a MOSSE filter for the difference between the horizon line and the water line in the image.
This results in an average position error of 3.0m, i.e. inferior to using merely the horizon line.
Our explanation is that for cameras at low height with small gracing angles to the sea, the water line position in the image is very sensitive to the actual height of the camera above the sea level.
Also, the variance of the water line position in the image is small unless the USV is very close to land.
These two factors make water line registration inaccurate for cameras on small USVs.





\section{Conclusions}
In this paper, we have investigated the problem of providing accurate position measurements for a USV operating in coastal areas or in the archipelago when GPS signals are denied or failing. We propose a position estimation method where the horizon line is extracted in a 360$^{\circ}$ panoramic image around the USV. We design two CNN architectures. The first CNN determines an approximate horizon line in the image and implicitly determines the camera orientation. The image is warped to compensate for the camera orientation. The second CNN  extracts the pixelwise horizon line and water line in the warped image. The extracted horizon line is correlated with DEM data in the Fourier domain using a MOSSE correlation filter. Finally, we determine the location of the maximum correlation score over the search area to estimate the position of the USV. Our approach provides promising results by achieving position estimates with GPS-level accuracy.

We have shown that the proposed method with registration of the horizon line in the Fourier domain outperforms a classical method with registration in the spatial domain, both in terms of position accuracy and computational speed. We found that for a small USV, with the camera nominally located only 1m above the sea level, registration of the water line with DEM data was not robust and degraded the position estimates obtained when using registration of the horizon line only. The observation that horizon line registration is sufficient to provide very accurate position estimates suggests that our proposed method can be generalized to other domains and also be employed by UGVs and UAVs operating on land.

\section*{Acknowledgment}
This work was partially supported by the Wallenberg AI, Autonomous Systems and Software Program (WASP) funded by the Knut and Alice Wallenberg Foundation.

This work was supported by the Swedish Foundation for Strategic Research (Smart Systems: RIT 15-0097).

This research is supported by  CENIIT grant (18.14), and VR starting grant (2016-05543).
%


\bibliographystyle{IEEEtran}
\bibliography{horizon_bib}

\end{document}